\documentclass[11pt]{article}

\usepackage[final]{acl}

\usepackage{times}
\usepackage{latexsym}

\usepackage[T1]{fontenc}

\usepackage[utf8]{inputenc}

\usepackage{microtype}

\usepackage{inconsolata}

\usepackage{graphicx}

\usepackage{algorithm}
\usepackage{algorithmic}
\usepackage{float}
\usepackage{placeins}
\usepackage{amsmath}

\usepackage{enumitem}
\usepackage{booktabs}   
\usepackage{colortbl}   
\usepackage{tabularx}   
\usepackage{multirow}   
\usepackage{caption}    

\title{ConvergeWriter: Data-Driven Bottom-Up Article Construction}

\author{
  \textbf{Binquan Ji}, \textbf{Jiaqi Wang}, \textbf{Ruiting Li},
  \textbf{Xingchen Han}, \textbf{Haibo Luo},\\
  \textbf{Yiyang Qi}, \textbf{Shichao Wang}, \textbf{Yifei Lu},
  \textbf{Yuantao Han}, \textbf{Feiliang Ren}\textsuperscript{*}\\
  School of Computer Science and Engineering,\\
  Northeastern University, Shenyang 110819, China\\
  \texttt{jibinquan@foxmail.com, renfeiliang@cse.neu.edu.cn}
}

\begin{document}
\maketitle
\begingroup
\renewcommand{\thefootnote}{*}
\footnotetext{Corresponding author.}
\endgroup
\begin{abstract}
Large Language Models (LLMs) excel at text generation, but still struggle to produce accurate, long-form documents grounded in large knowledge bases. Existing ``top-down'' approaches---which plan before retrieving evidence---often result in fragmented or unfaithful content due to misalignment between the plan and available knowledge.

To overcome this, we introduce a ``bottom-up'' framework based on a ``Retrieval-First for Knowledge, Clustering for Structure'' strategy. Our method begins by establishing the ``knowledge boundaries'' of the source corpus through comprehensive retrieval, followed by unsupervised clustering to group evidence into distinct ``knowledge clusters.'' These clusters objectively inform a hierarchical outline and subsequent text generation, ensuring strict adherence to the source and significantly reducing hallucination.

Experimental results on both 14B- and 32B-parameter models demonstrate that our method achieves performance comparable to or exceeding state-of-the-art baselines, and is expected to demonstrate unique advantages in knowledge-constrained scenarios that demand high fidelity and structural coherence. Our work presents an effective paradigm for generating reliable, structured, long-form documents, paving the way for more robust LLM applications in high-stakes, knowledge-intensive domains.
\end{abstract}

\section{Introduction}

\label{sec:introduction}

\begin{figure}[t]
\centering
\includegraphics[width=0.9\columnwidth]{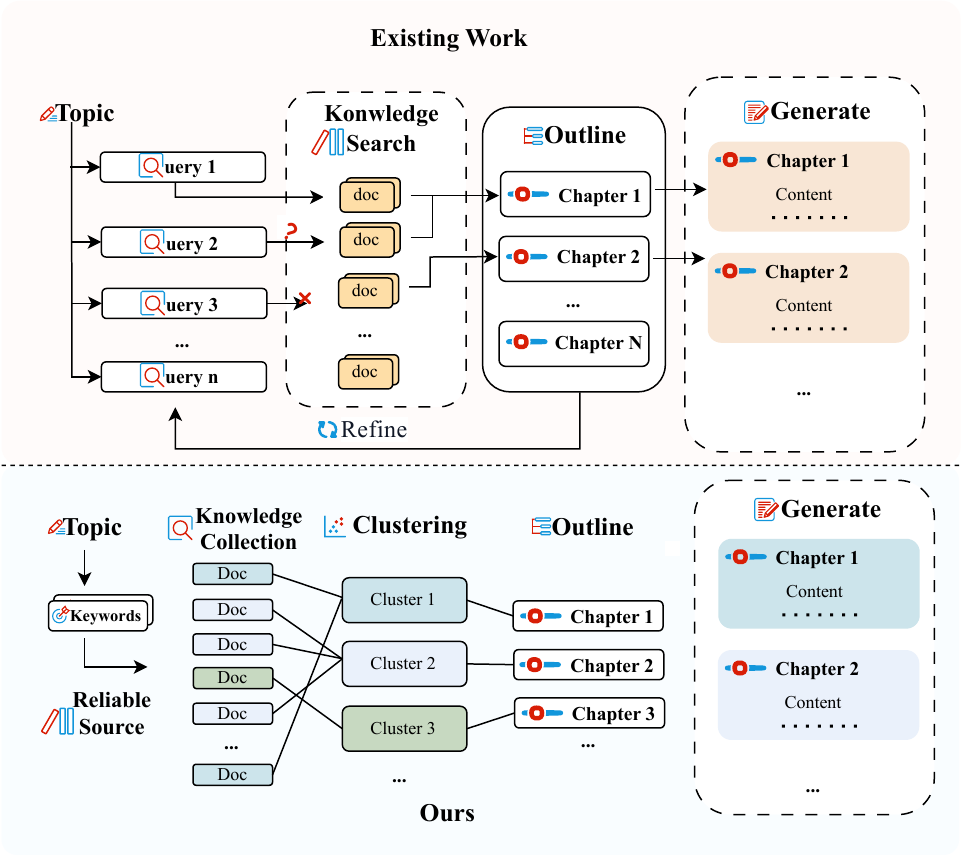} 
\caption{Comparison of article-construction pipelines. Top-down methods generate an outline or questions from the input topic and then retrieve evidence for the predefined structure. ConvergeWriter instead retrieves and clusters evidence to estimate the corpus's knowledge boundary, induces an outline from the resulting knowledge clusters, and generates sections grounded in those clusters.}
\label{fig:top_down_vs_ours}
\end{figure}

In recent years, Large Language Models (LLMs) have made significant advances in Natural Language Processing (NLP), excelling in tasks like open-domain QA and text generation~\cite{yang2023docimprovinglongstory, zhao2025surveylargelanguagemodels, yue2025surveylargelanguagemodel}. However, effectively incorporating external knowledge to produce detailed, credible, and hallucination-free long-form documents remains challenging. This task requires not only coherent long-text generation but also integration of reference materials beyond the model's context window, while ensuring accuracy and structural integrity~\cite{xu2025comprehensivesurveydeepresearch}.

Specifically, leveraging external knowledge for long-text generation faces three core challenges:

\begin{itemize}
    \item \textbf{Massive Context Processing:} Referencing large volumes of background material demands processing ultra-long texts, posing computational challenges and requiring efficient extraction of salient information~\cite{liu-etal-2024-lost, NEURIPS2024_d07a9fc7}.
    
    \item \textbf{Knowledge Organization and Structuring:} Multi-source knowledge must be structured logically to avoid redundancy and ensure coherence, going beyond mere information listing~\cite{skarlinski2024languageagentsachievesuperhuman}.
    
    \item \textbf{High-Fidelity and Low-Hallucination:} In accuracy-critical domains (e.g., finance, healthcare), generated text must strictly adhere to credible sources under ``closed knowledge'' constraints to eliminate hallucinations~\cite{10.1145/3571730, shao-etal-2024-assisting}.
\end{itemize}

Existing methods like OmniThink~\cite{xi2025omnithinkexpandingknowledgeboundaries} and STORM~\cite{shao-etal-2024-assisting} typically adopt a top-down, hypothesis-driven strategy: first outlining a structure or raising questions, then retrieving evidence to fit this framework. This approach, however, may misalign with the knowledge base, leading to inefficient retrieval, fragmented content, and reduced credibility~\cite{shi-etal-2024-replug}.

To overcome these issues, we propose \textbf{ConvergeWriter}, a bottom-up, data-driven framework for long-text generation based on ``retrieval-first for knowledge, clustering for structure.'' It begins by retrieving relevant knowledge, then plans the article's content and structure based on the actual evidence, constraining generation to available knowledge and avoiding LLM hallucinations. Our approach addresses each challenge as follows:

\begin{itemize}
    \item \textbf{For Massive Context Processing:} We use multi-round relevance-expanding retrieval to gather documents, then apply tree-structured hierarchical summarization (inspired by RAPTOR~\cite{sarthi2024raptor}) to compress text while preserving core information, significantly increasing processable document length and volume.
    
    \item \textbf{For Knowledge Organization and Structuring:} Retrieved documents are clustered into traceable \textbf{Knowledge Clusters}, forming a content-grounded framework. The article outline is derived directly from these clusters, ensuring structural coherence and knowledge-base alignment.
    
    \item \textbf{For High-Fidelity and Low-Hallucination:} The outline is fully data-driven. During paragraph generation, we use an ``Outline--Knowledge Cluster Jointly Guided'' retrieval-augmented approach, anchoring each segment to verifiable documents, mitigating hallucinations caused by information gaps or model fabrication.
\end{itemize}

Experiments using Wikipedia as the sole retrieval source show that ConvergeWriter achieves competitive or superior performance compared to state-of-the-art baselines on both 14B- and 32B-parameter models, validating its effectiveness in knowledge-bound and high-reliability scenarios.

\begin{figure*}[t]
\centering
\includegraphics[width=0.95\textwidth]{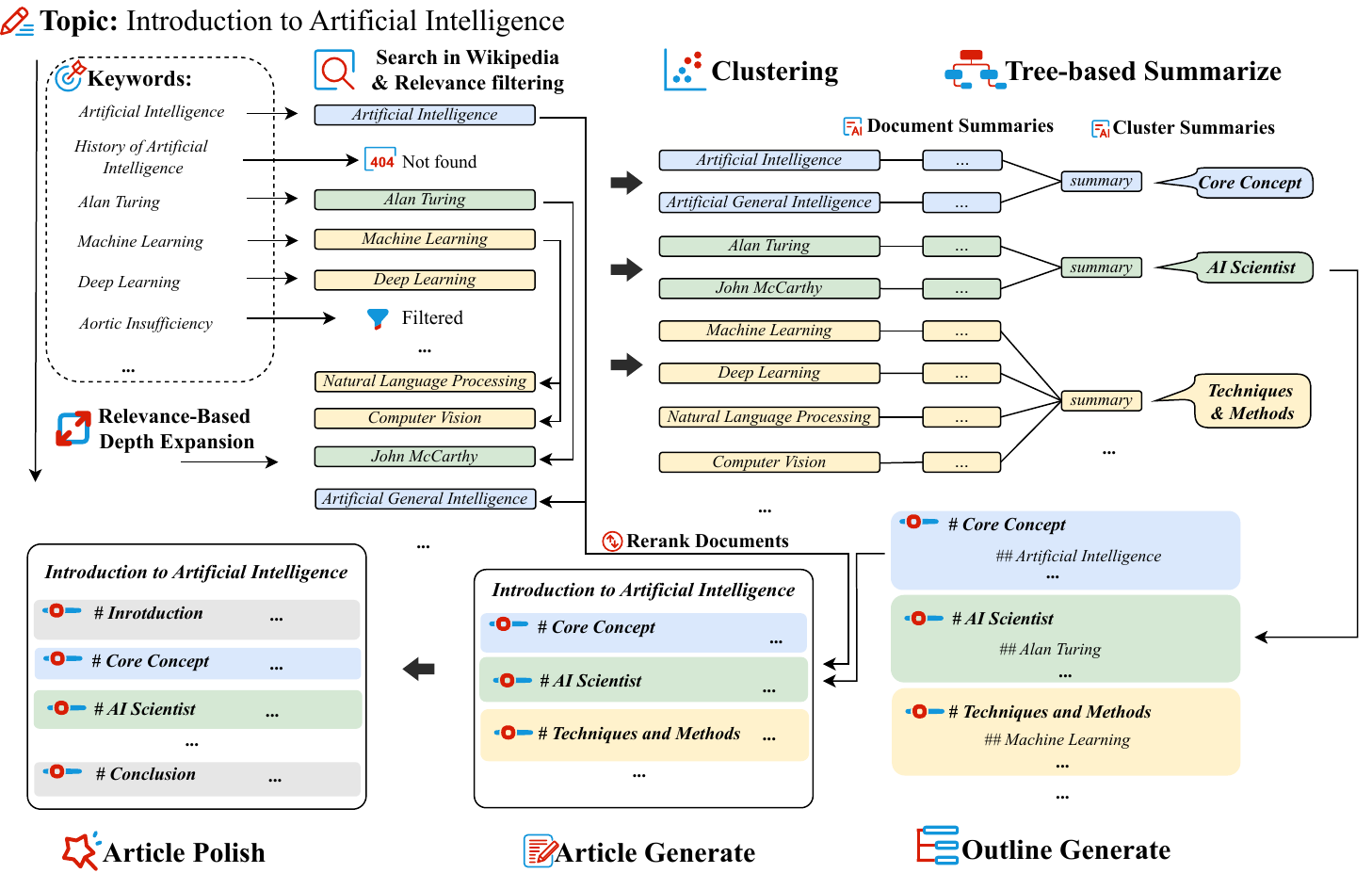} 
\caption{ConvergeWriter's workflow: (1) iterative relevance-expanding knowledge retrieval and document screening; (2) unsupervised clustering with tree-structured summarization; (3) LLM-based outline generation; and (4) recall-enhanced chapter-wise content generation, introduction/conclusion supplementation, and polishing.}
\label{fig:workflow}
\end{figure*}

\section{Related Work}
\label{sec:related_work}

Recent advancements in long-text generation with Large Language Models largely follow a hypothesis-driven, top-down paradigm. STORM~\cite{shao-etal-2024-assisting} decomposes writing into iterative stages, including multi-perspective knowledge gathering, outlining, and drafting. Similarly, OmniThink~\cite{xi2025omnithinkexpandingknowledgeboundaries} expands knowledge via dynamic ``information tree'' construction and abstraction, while LongWriter~\cite{bai2025longwriter} separates generation into planning and sequential paragraph writing. Across these approaches, retrieval mainly validates or supplements a predefined outline constructed before the full knowledge scope is known~\cite{bai2024longwriterunleashing10000word}.

Multi-agent frameworks further extend this paradigm. AutoPatent~\cite{wang2024autopatentmultiagentframeworkautomatic} employs Planner/Author/Reviewer agents guided by a predefined ``PGTree,'' using Reference-Reviewer-Augmented Generation (RRAG) for local optimization. The cognitive writing framework of \citet{DBLP:journals/corr/abs-2502-12568} simulates human writing through hierarchical planning and parallel generation with iterative revision. WebThinker~\cite{li2025webthinkerempoweringlargereasoning} integrates real-time web exploration in a ``Think-Search-Draft'' loop with reinforcement learning. Despite improved task decomposition and coherence, these methods still rely on an initial plan formed without verified knowledge availability, risking structural misalignment with retrievable evidence.

Despite their effectiveness, existing approaches face three challenges in knowledge-constrained scenarios:

\begin{enumerate}
    \item \textbf{Knowledge-Structure Mismatch:} Predefined outlines may exceed retrievable evidence, causing fragmented or fabricated sections.
    
    \item \textbf{Static Planning:} Outlines cannot adapt to actual knowledge coverage, forcing models to speculate.
    
    \item \textbf{Opaque Provenance:} The link between generated content and source documents remains indirect, limiting verifiability.
\end{enumerate}

To address these issues, ConvergeWriter proposes a data-driven, bottom-up framework. It first performs retrieval-first clustering to identify knowledge boundaries, then induces the article structure accordingly. This grounds generation in available evidence, mitigating outline hallucination while improving traceability through cluster-anchored generation.

\section{Methodology}
\label{sec:Methodology}

The core workflow of ConvergeWriter is illustrated in Figure~\ref{fig:workflow}. The method begins with multiple rounds of relevance-expanding knowledge retrieval and supporting document screening. Subsequently, unsupervised clustering is performed on the documents to form knowledge clusters, combined with tree-structured summarization to construct a hierarchical knowledge structure. Following this, a logically coherent outline (where main chapters correspond to knowledge clusters) is generated by an LLM based on the integrated summaries. Finally, a recall-enhanced generation paradigm jointly driven by the outline and knowledge clusters is employed for chapter-wise content generation and integration. Content for each chapter is modularly generated by the LLM driven by the outline and constructed knowledge clusters, followed by the supplementation of the introduction and conclusion sections, along with comprehensive article polishing.

\subsection{Relevance-Expanding Knowledge Retrieval}
\label{subsec:retrieval}

To enhance the concreteness and depth of the generated content, we design a relevance-expanding knowledge retrieval module that systematically collects high-coverage documents highly relevant to the topic $T$ from external knowledge bases (e.g., Wikipedia). This process is completed through an iterative two-stage approach:

\textbf{Stage 1: Breadth-First Retrieval}\\
First, we use a Large Language Model (LLM) $\mathcal{M}$ to generate an initial keyword set based on topic $T$. Here and throughout, $\mathcal{I}_{\cdot}$ denotes the prompt or instruction template used by the corresponding LLM module:
\begin{equation}
    \mathcal{K}_0 = \mathcal{M}(T; \mathcal{I}_{\text{keyword-gen}})
\end{equation}
Then invoke the retrieval interface $\mathcal{R}$ (e.g., Wikipedia API) using $\mathcal{K}_0$ to obtain the preliminary document set $\mathcal{D}^{(0)} = \mathcal{R}(\mathcal{K}_0)$. Subsequently, perform relevance filtering on $\mathcal{D}^{(0)}$ via the LLM, retaining only documents relevant to $T$ to form a refined set:

\begin{equation}
\begin{aligned}
\mathcal{D}^{(1)} =
\{ d \in \mathcal{D}^{(0)} \mid\;
& \mathcal{M}(T, d; \mathcal{I}_{\text{rel-filter}}) \\
&= \text{Relevant} \}
\end{aligned}
\end{equation}

\textbf{Stage 2: Relevance-Based Depth Expansion}\\
To deepen the knowledge collection and mitigate potential omissions of critical subtopics or details, we perform relevance-based depth expansion based on $\mathcal{D}^{(1)}$. For each document $d$ in $\mathcal{D}^{(1)}$, utilize the LLM $\mathcal{M}$ to generate expanded keywords guided by both $T$ and the content of $d$, prompting it to explore deeper potential information within the context of the current topic, thereby forming an expanded keyword set:
\begin{equation}
    \mathcal{K}_{\text{ext}}= \bigcup_{d \in \mathcal{D}^{(1)}} \mathcal{M}(T,d;\mathcal{I}_{\text{depth-exp}})
\end{equation}
Use $\mathcal{K}_{\text{ext}}$ for secondary retrieval to obtain $\mathcal{D}^{(2)}_{\text{raw}}$, followed by the same relevance filtering as in Stage 1. Finally, merge the results from both stages to construct the high-quality knowledge document set:
\begin{equation}
\begin{aligned}
\mathcal{D}^{*} =
\mathcal{D}^{(1)} \cup
\{ d \in \mathcal{D}^{(2)}_{\text{raw}}
\mid\;
& \mathcal{M}(T, d; \mathcal{I}_{\text{rel-filter}}) \\
& = \text{Relevant} \}
\end{aligned}
\end{equation}

Through this iterative two-stage retrieval process, we construct a high-quality, highly relevant, and sufficiently deep knowledge document set $\mathcal{D}^{*}$ to support subsequent article generation. This retrieval-first strategy ensures that subsequent content planning and generation strictly adhere to the boundaries of practically available knowledge evidence.

\subsection{Knowledge Structurization}
\label{subsec:knowledge_structurization}

After obtaining the high-quality candidate document set $\mathcal{D}^{*}$, we organize it structurally to reveal its inherent topical distribution, paving the way for generating a logically coherent article outline in subsequent stages. The core of this process is transforming the unstructured document collection into a set of semantically meaningful knowledge clusters.

Specifically, this step consists of two phases: document clustering and hierarchical summarization:

\textbf{1. Document Clustering via Silhouette Coefficient Optimization}
First, we employ a pre-trained embedding model $\mathcal{E}$ to map each document $d_i$ in $\mathcal{D}^{*}$ to a semantic vector $v_i = \mathcal{E}(d_i)$, projecting the entire document set into a high-dimensional vector space.

Next, to partition these vectors into semantically cohesive topic clusters, we utilize the $k$-means clustering algorithm. To automatically determine the optimal number of clusters $k$, we adopt the silhouette coefficient~\cite{ROUSSEEUW198753} as the evaluation metric. We iterate over $k$ values within a predefined range $[k_{\min}, k_{\max}]$ and select the $k$ that maximizes the average silhouette coefficient $\bar{s}(k)$:
\begin{equation}
k^* = \underset{k \in [k_{\min}, k_{\max}]}{\arg\max}  \bar{s}(k)
\end{equation}
To enhance computational efficiency for large document sets, we approximate the silhouette coefficient calculation through random sampling. Finally, using the determined optimal $k^*$, we execute the $k$-means algorithm on all document embedding vectors, obtaining a collection of $k^*$ knowledge clusters $\{C_1, C_2, \ldots, C_{k^*}\}$.

\textbf{2. Tree-Structured Summarization for Long Contexts}
After clustering, each cluster $C_j$ may contain numerous documents. Directly concatenating document contents could easily exceed the model's context window limit. To address this, we adopt a hierarchical tree-structured summarization strategy:

\begin{itemize}
    \item \textbf{Leaf Node Summarization:} For each document $d_i$ in cluster $C_j$, we invoke an LLM to generate a concise summary:
    \begin{equation}
    s_i = \mathcal{M}(d_i; \mathcal{I}_{\text{summarize}}).
    \end{equation}   
    \item \textbf{Root Node Summarization:} We concatenate all document summaries within the cluster $\{s_i\}_{d_i \in C_j}$ and input them into the LLM again to generate a cluster-level descriptive summary:
    \begin{equation}
    S_j = \mathcal{M}\left( \texttt{concat}(\{s_i\}); \mathcal{I}_{\text{cluster-summarize}} \right)
    \end{equation}
\end{itemize}

Through this process, the original document set $\mathcal{D}^{*}$ is transformed into a collection of semantically cohesive knowledge clusters $\{C_j\}_{j=1}^{k^*}$, each accompanied by a refined summary $S_j$. This hierarchical knowledge representation not only reveals the intrinsic structure of the available knowledge but also provides direct and reliable input for generating a high-quality article outline based on this structure in the subsequent stage.

\subsection{Structured Knowledge Mapping-Based Outline Generation}
After completing the previous stage, a set of highly cohesive knowledge clusters $\{C_j\}_{j=1}^{k^*}$ is obtained, where each cluster $C_j$ encapsulates the core information regarding a specific dimension of topic $T$ in the knowledge base, characterized by its integrated summary $S_j$. The objective of this stage is to leverage these structured knowledge summaries to generate a logically rigorous and focus-oriented outline $O$ for an article on topic $T$.

\begin{equation}
O = \mathcal{M}\left( \{S_j\}_{j=1}^{k^*}; \mathcal{I}_{\text{outline\_gen}} \right)
\end{equation}

To ensure the outline strictly adheres to the actual content boundaries of the knowledge base and mitigates hallucination risks, the outline generation process is subject to dual constraints: the model must autonomously optimize the sequence of sections to form a cognitively logical argument flow, while exclusively extracting highly relevant core information for topic $T$ from the summary set $\{S_j\}_{j=1}^{k^*}$. Furthermore, to guarantee machine readability, format consistency, and provide explicit guidance for subsequent generation, formatting constraints are imposed on the large language model (LLM) output. Specifically, $O$ must strictly comply with Markdown syntax, and with the exception of the ``Introduction'' and ``Conclusion'' sections, each main section $\text{Sec}_i$ must explicitly and uniquely map to a knowledge cluster $C_j$, i.e.:


\begin{equation}
\forall \mathrm{Sec}_i \in O_{\text{body}},
\exists!\, C_j \in \mathcal{C}
\;:\;
f(\mathrm{Sec}_i) = C_j
\end{equation}

The core design principle ensures the \textbf{verifiability} and \textbf{source determinacy} of the article's central arguments, effectively circumventing hallucination risks.

\subsection{Outline--Knowledge Cluster Joint-Driven Section-Wise Retrieval for Article Generation}
This stage constructs the complete article based on the generated outline $O$. First, $O$ is parsed into discrete section units $\{\text{Sec}_1, \text{Sec}_2, \dots, \text{Sec}_K\}$. For each main section $\text{Sec}_i$ ($i \notin \{1, K\}$), relevant documents from its corresponding knowledge cluster $C_j$ are retrieved, and a Ranker model re-ranks them to produce an enhanced document set $\mathcal{D}_{\text{sec}_i}^*$. The section content is then generated as:

\begin{equation}
\text{Sec}_i^* = \mathcal{M}(\text{Sec}_i, \mathcal{D}_{\text{sec}_i}^*; \mathcal{I}_{\text{section\_gen}})
\end{equation}

After generating all main sections $\text{Sec}_i^*$ independently, we concatenate them to form the article body $\mathcal{A}_{\text{draft}}$. We then use the topic and $\mathcal{A}_{\text{draft}}$ to generate the ``Introduction'' section $\text{Sec}_1^*$ and ``Conclusion'' section $\text{Sec}_K^*$ separately. Combining these sections with the article body yields the complete draft $\mathcal{A}_{\text{full\_draft}}$.

Finally, the large language model $\mathcal{M}$ performs global polishing on $\mathcal{A}_{\text{full\_draft}}$ to produce a grammatically fluent, stylistically consistent, and coherent final article:

\begin{equation}
\mathcal{A}_{\text{final}} = \mathcal{M}(\mathcal{A}_{\text{full\_draft}}; \mathcal{I}_{\text{refine}})
\end{equation}

\section{Experiments}
To comprehensively evaluate the effectiveness of \textbf{ConvergeWriter}, we follow the experimental paradigm of prior work (e.g., OmniThink~\cite{xi2025omnithinkexpandingknowledgeboundaries}) and conduct experiments on a fixed subset of topics from the \textbf{WildSeek} dataset~\cite{jiang-etal-2024-unknown}. WildSeek is derived from real users' multi-turn information-exploration sessions with the open-web applications STORM~\cite{shao-etal-2024-assisting} and Co-STORM~\cite{jiang-etal-2024-unknown}, and is designed to evaluate complex information-retrieval and knowledge-consolidation systems. Our experiments focus on a core task: given a topic title and a single designated knowledge source (i.e., the Wikipedia retrieval API), assess the capability of different methods to collect and organize relevant information and generate well-structured, informative articles. This task setup simulates applications in vertical domains (e.g., finance and scientific report writing) with well-defined knowledge boundaries and high information-credibility requirements. Details of the dataset subset and evaluation protocol are provided in Appendix~\ref{sec:evaluation_protocol}.

\subsection{Evaluation Metrics}
To comprehensively and multi-dimensionally evaluate the performance of different methods in knowledge organization and article generation, and to address the core challenges outlined in the introduction (massive context processing, knowledge structuring, high trustworthiness), we employ the following four complementary evaluation metrics:

\begin{enumerate}
    \item \textbf{Average Article Length (Length)}: Records the average whitespace-delimited word count of generated articles within the fixed context-length limit of the LLM. This metric directly measures the textual output volume achievable by a method under the given constraint.
    \item \textbf{Cited Documents (Cited Docs)}: Counts the average number of explicitly cited supporting documents per article, measuring the breadth and depth of evidence collection and application.
    \item \textbf{LLM Automatic Evaluation}: Uses \textbf{Qwen3-32B}~\cite{yang2025qwen3technicalreport} as an evaluator to automatically score articles generated by all methods (on a 0--5 scale) across four key quality dimensions:
    \begin{itemize}
        \item \textbf{Relevance}: Assesses how closely the article content aligns with the initial topic intent, avoiding the introduction of irrelevant content or digressions.
        \item \textbf{Breadth}: Measures the comprehensiveness of an article by evaluating how broadly it covers the core aspects of the topic.
        \item \textbf{Depth}: Gauges the extent to which the article provides in-depth, detailed analysis of the topic and related domains, surpassing superficial descriptions.
        \item \textbf{Novelty}: Evaluates the article's capability to introduce relevant, non-directly derived new insights or perspectives beyond the user's initial intent.
    \end{itemize}
We assess the validity of this automatic evaluation method through two complementary consistency studies: a parallel evaluation using DeepSeek-v4-flash as a heterogeneous LLM judge, and a comparison between LLM-based evaluation and human judgments. Detailed experimental procedures and results are provided in Appendices~\ref{sec:heterogeneous_llm_judge} and~\ref{sec:llm_human_consistency}.
    \item \textbf{Document Coverage (Coverage \%)}: This metric quantifies the extent to which the content of the generated article is actually supported by the knowledge repository, directly addressing the challenge of high fidelity and low hallucination. The calculation procedure is as follows:
    \begin{enumerate}[label=\roman*)]
        \item Segment the article into paragraph-level verification units using the deterministic preprocessing rules in Appendix~\ref{sec:evaluation_protocol}.
        \item For each paragraph, retrieve the two most similar documents from the document collection in the knowledge base; an evaluation of the number of retrieved documents is provided in Appendix~\ref{sec:coverage_sensitivity}.
        \item Prompt the LLM to strictly judge whether the retrieved documents support the paragraph's description (i.e., content is relevant and conflict-free).
        \item If at least one document is judged to support the paragraph, the paragraph is considered supported; otherwise, it is considered unsupported.
        \item Calculate the percentage of supported paragraphs relative to the total number of paragraphs in the article.
    \end{enumerate}
\end{enumerate}

\begin{table*}[t]
\centering
\small
\renewcommand{\arraystretch}{1.15}
\setlength{\tabcolsep}{4pt}
\begin{tabular}{@{}lrrrrrrrr@{}}
\toprule
\multirow{2}{*}{\textbf{Method}} &
\multirow{2}{*}{\textbf{Length}} &
\multirow{2}{*}{\textbf{\shortstack{Cited\\Docs}}} &
\multicolumn{5}{c}{\textbf{Rubric Grading}} &
\multirow{2}{*}{\textbf{Coverage \%}} \\
\cmidrule(lr){4-8}
 & & &
 \textbf{Relevance} & \textbf{Breadth} & \textbf{Depth} & \textbf{Novelty} & \textbf{AVG} & \\
\midrule
\multicolumn{9}{c}{\textbf{Qwen3-14B without Reasoning}} \\
\midrule 
Direct RAG & 4,848.95 & 4.27 & 3.92 & 3.90 & 3.79 & 2.81 & 3.60 & 36.67 \\
Two-Stage RAG & \textbf{54,744.76} & 3.63 & 3.77 & 4.22 & \underline{4.39} & \underline{3.98} & 4.14 & 7.12 \\
STORM & 14,732.00 & 4.97 & \underline{4.80} & \underline{4.58} & 4.35 & 2.92 & \underline{4.16} & 24.91 \\
OmniThink & \underline{27,376.73} & \underline{6.03} & 4.43 & 4.12 & 3.79 & 3.35 & 3.92 & \underline{53.88} \\
ConvergeWriter (ours) & 25,913.52 & \textbf{9.08} & \textbf{4.93} & \textbf{4.95} & \textbf{4.97} & \textbf{4.22} & \textbf{4.77} & \textbf{80.14} \\
\midrule
\multicolumn{9}{c}{\textbf{Qwen3-32B with Reasoning}} \\
\midrule 
Direct RAG & 9,252.43 & \underline{8.67} & \underline{4.93} & \underline{4.93} & \underline{4.92} & 3.72 & \underline{4.62} & \underline{61.96} \\
Two-Stage RAG & \textbf{68,524.68} & 3.54 & 3.54 & 4.46 & 4.54 & 4.08 & 4.16 & 2.08 \\
STORM & 24,105.05 & 7.95 & 4.78 & 4.75 & 4.75 & 3.85 & 4.53 & 35.84 \\
OmniThink & \underline{35,917.12} & \textbf{9.52} & 4.43 & 4.38 & 4.37 & \underline{4.13} & 4.33 & 43.22 \\
ConvergeWriter (ours) & 22,629.35 & 7.98 & \textbf{4.97} & \textbf{4.95} & \textbf{4.93} & \textbf{4.58} & \textbf{4.86} & \textbf{70.51} \\
\bottomrule
\end{tabular}
\caption{Evaluation results on the WildSeek dataset. Length is the average whitespace-delimited word count; Cited Docs is the average number of cited documents; Rubric Grading reports four LLM-evaluated dimensions and their average; and Coverage is the percentage of supported paragraphs. Bold and underlined values denote the best and second-best results within each model setting, respectively.}
\label{tab:results}
\end{table*}

\subsection{Baseline Models}
To accurately evaluate ConvergeWriter's performance, we rigorously follow the incremental comparison principle, selecting representative Retrieval-Augmented Generation (RAG) baselines for comparison:
\begin{itemize}
    \item \textbf{Direct RAG}~\cite{NEURIPS2020_6b493230}: The foundational paradigm retrieves relevant documents once based on the topic and directly inputs them to the LLM for single-step generation of the complete article.
    \item \textbf{Two-Stage RAG}: A two-stage approach: First, it generates a retrieval-augmented outline; then, it expands content section by section based on the outline (retrieving topic-relevant documents again for each section); finally, it concatenates the sections into a full article.
    \item \textbf{STORM}~\cite{shao-etal-2024-assisting}: It delves into topic details through simulated multi-perspective dialogues and systematically integrates information to produce a structured outline. The article is subsequently written section by section according to this outline.
    \item \textbf{OmniThink}~\cite{xi2025omnithinkexpandingknowledgeboundaries}: It extends knowledge breadth via an ``information tree,'' deepens cognitive associations through a ``concept pool,'' and dynamically broadens knowledge boundaries to accomplish concept abstraction, outline generation, and citation-aware paragraph writing.
\end{itemize}
These baselines span from basic retrieval augmentation (\textbf{Direct RAG}~\cite{NEURIPS2020_6b493230}) and structured staged augmentation (\textbf{Two-Stage RAG}) to state-of-the-art complex knowledge planning and integration methods (\textbf{STORM}~\cite{shao-etal-2024-assisting}, \textbf{OmniThink}~\cite{xi2025omnithinkexpandingknowledgeboundaries}), aiming to isolate and assess the impact of diverse knowledge utilization strategies and content organization structures on long-text generation.

\subsection{Experimental Details}
To systematically evaluate the generalization performance of \textbf{ConvergeWriter} across varying language model scales and reasoning capabilities, we conduct comprehensive experiments on two representative model configurations:
\begin{enumerate}
    \item \textbf{Qwen3-14B (Basic Reasoning Configuration)}~\cite{yang2025qwen3technicalreport}: We disable the model's deep thinking mode to assess method performance under the model's fundamental capabilities.
    \item \textbf{Qwen3-32B (Enhanced Reasoning Configuration)}~\cite{yang2025qwen3technicalreport}: We enable the model's deep thinking mode to verify whether methods can effectively harness and guide the enhanced reasoning capabilities of more powerful models to yield superior results.
\end{enumerate}
All experiments enforce a maximum context-length limit of 24K tokens. For baseline methods involving multi-round interactions (e.g., \textbf{STORM}~\cite{shao-etal-2024-assisting} and \textbf{OmniThink}~\cite{xi2025omnithinkexpandingknowledgeboundaries}), the per-round context length is capped at 12K tokens to prevent overflow and ensure fair comparison. Key experimental settings include:
\begin{itemize}
    \item \textbf{Knowledge Source and Retrieval}: We uniformly employ the Wikipedia official retrieval API as the external knowledge source. All retrieval-augmented methods (including \textbf{ConvergeWriter} and RAG baselines) acquire knowledge through this API.
    \item \textbf{Embedding and Reranking Models}: The embedding model (Qwen3-Embedding-0.6B~\cite{zhang2025qwen3embeddingadvancingtext}) and reranker model (Qwen3-Reranker-0.6B~\cite{zhang2025qwen3embeddingadvancingtext}) are consistently used across all methods to maintain retrieval-process integrity.
    \item \textbf{Computational Resources}: All experiments are executed on three NVIDIA A6000 GPUs.
\end{itemize}

\subsection{Experimental Results}
\label{sec:experimental_results}

Our experimental results are detailed in Table~\ref{tab:results}. The results demonstrate that ConvergeWriter exhibits significant advantages across core metrics, validating the effectiveness of its design principles. Specifically:

\textbf{(1) Bottom-Up Data-Driven Strategy Significantly Enhances Verifiability and Credibility}\\
The Document Coverage of ConvergeWriter on the 14B and 32B models reaches 80.14\% and 70.51\%, respectively, significantly outperforming the baselines. This confirms that its core strategy---``retrieval-first for knowledge, clustering for structure''---can a priori build a traceable knowledge cluster framework. This helps anchor the generated content in retrieved evidence, reducing unsupported content while enhancing credibility and verifiability. In contrast, traditional ``top-down'' approaches, due to their inherently ``hypothesis-driven'' nature, often lead to conceptualization that diverges from the actual content of the knowledge base, resulting in generated content lacking reliable source support. Additionally, we evaluate claim-level evidence support across all five methods. ConvergeWriter continues to outperform all baselines in terms of claim support rate, further supporting the conclusion above. The complete evaluation protocol and results are reported in Appendix~\ref{sec:claim_level_support}.

\textbf{(2) Objective Generation Based on Clustered Knowledge Structures Improves Overall Article Quality}\\
Compared to other structure-oriented methods like STORM~\cite{shao-etal-2024-assisting} and OmniThink~\cite{xi2025omnithinkexpandingknowledgeboundaries}, ConvergeWriter's core advantage lies in its article structure being driven by objective data (derived from document clustering results) rather than subjective model conceptualization. This enables the article's breadth and depth to more authentically reflect the full spectrum of the knowledge base. Notably, ConvergeWriter also achieved the highest scores on the Novelty metric (4.22 and 4.58). This indicates that by clustering documents, the model can uncover latent, non-intuitive connections between knowledge points, generating more insightful perspectives rather than merely combining information.

\textbf{(3) Relevance-Expanding Knowledge Retrieval Optimizes Citation Breadth and Content Focus}\\
On the 14B model, the number of Cited Docs for ConvergeWriter reaches 9.08, demonstrating that its iterative retrieval and clustering mechanism effectively broadens the scope of knowledge exploration. Simultaneously, it generates articles of moderate length, achieving content conciseness while ensuring high quality and high credibility. In contrast, although the Two-Stage RAG method produces extremely long text, its low Support Document Coverage and mediocre ratings indicate substantial content redundancy and deviation from knowledge sources, resulting in ineffective information.


\textbf{(4) Framework Design Confers Strong Model Generalization Capability}\\
Experimental results exhibit consistent advantages across model scales (14B vs. 32B) and inference modes (standard vs. Deep Thinking Mode). Although performance generally improves with the more capable 32B model (with Deep Thinking Mode enabled), ConvergeWriter maintains its performance lead. This confirms that its success stems from the inherent advantages of the framework design, rather than reliance on specific model capabilities, demonstrating strong generalization and portability.

Additionally, we conduct a case study of the outputs generated by ConvergeWriter and STORM, as well as topic-level paired significance tests on the LLM-based evaluation results described above; detailed results are provided in Appendices~\ref{sec:qualitative_case_study} and~\ref{sec:paired_significance}, respectively.

\begin{table*}[!t]
\centering
\small
\setlength{\tabcolsep}{4pt}
\renewcommand{\arraystretch}{1.12}
\begin{tabular*}{\textwidth}{@{\extracolsep{\fill}}lrrrrr}
\toprule
\textbf{Method} & \textbf{Wall time (s)} & \textbf{Article words} & \textbf{Model tokens} & \textbf{s/1k words} & \textbf{tokens/1k words} \\
\midrule
ConvergeWriter & 289.1 & 3,323.1 & 66,446.6 & 92.7 & 21,113.5 \\
Direct RAG & 65.7 & 1,413.7 & 8,302.1 & 48.1 & 6,126.1 \\
Two-Stage RAG & 269.4 & 10,768.6 & 25,339.9 & 25.2 & 2,363.0 \\
STORM & 139.5 & 4,176.0 & 64,995.4 & 34.9 & 16,222.2 \\
OmniThink & 343.8 & 3,310.6 & 195,347.2 & 181.0 & 105,060.6 \\
\bottomrule
\end{tabular*}
\caption{Mean latency, output length, and model-token usage on the auxiliary 10-topic efficiency benchmark. Per-1,000-word metrics are computed for each topic and then averaged.}
\label{tab:efficiency_analysis}
\end{table*}

\subsection{Further Analysis}
\label{sec:further_analysis}
\subsubsection{Clustering Ablation}
\begin{figure}[htbp]
\centering
\includegraphics[width=0.9\columnwidth]{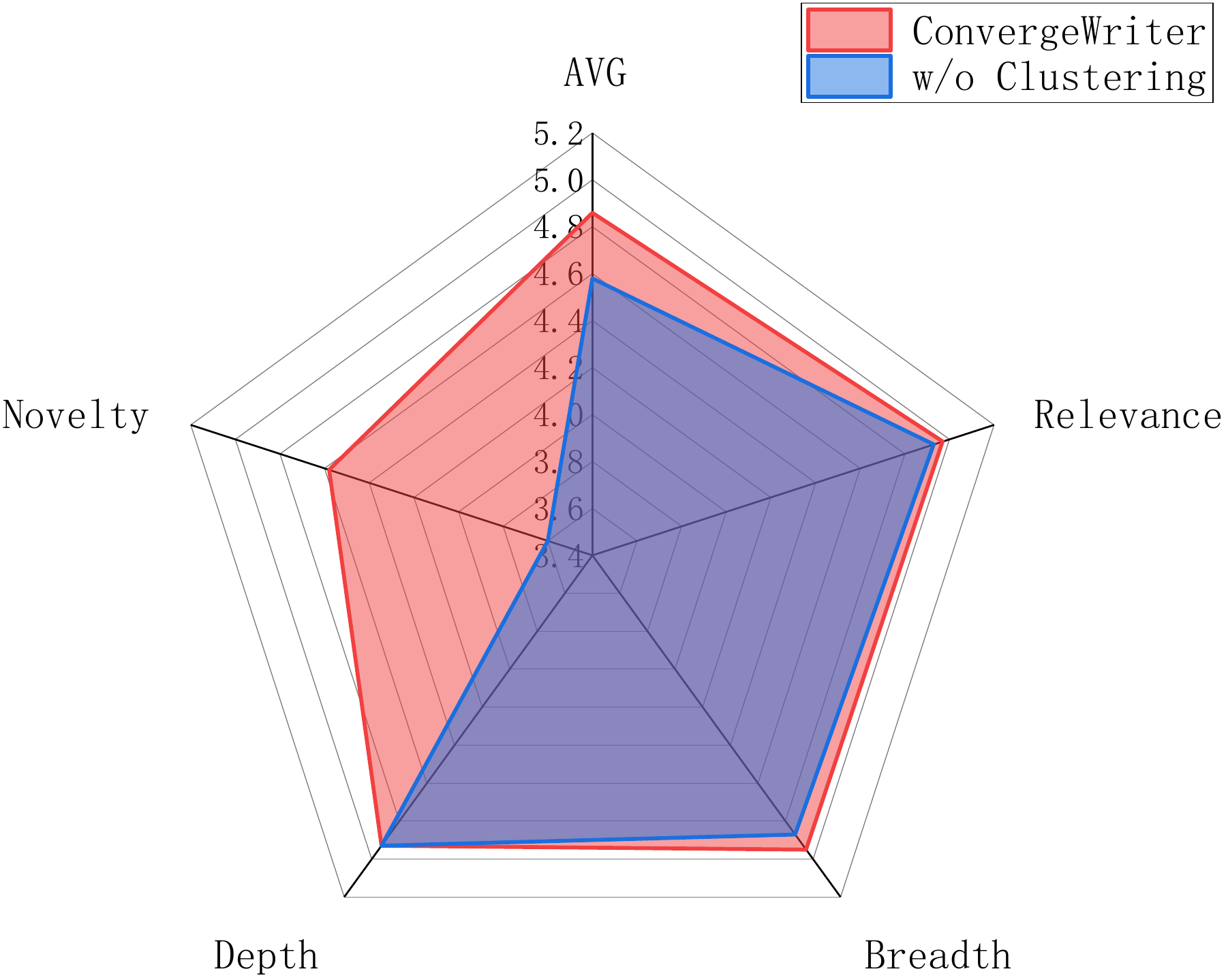} 
\caption{Clustering ablation. Sequential splitting fragments knowledge organization, causing a substantial novelty decline (4.58~$\rightarrow$~3.60) and less coherent output.}
\label{fig:clustering_ablation}
\end{figure}

To further examine the effectiveness of clustering in constructing a knowledge framework, we conduct an ablation study. Specifically, we remove the clustering component from ConvergeWriter (denoted as ``w/o Clustering'') and instead divide retrieved documents sequentially into five equal parts, while keeping all other components unchanged. The results are shown in Figure~\ref{fig:clustering_ablation}.

As illustrated, removing clustering significantly degrades article quality, with the average score dropping from 4.86 to 4.58. The largest decline occurs in ``Novelty'', decreasing from 4.58 to 3.60. We attribute this to the lack of effective knowledge organization. ConvergeWriter groups unstructured documents into semantically coherent clusters with high cohesion and low coupling, providing a focused basis for outline generation and ensuring thematic consistency. In contrast, the sequential partitioning in ``w/o Clustering'' is arbitrary and disrupts semantic relationships among documents, leading to disorganized knowledge and forcing the model to concatenate weakly related content, which harms coherence and novelty. 

A similar trend is observed in ``Breadth'' (4.95~$\rightarrow$~4.87). Clustering helps identify diverse thematic dimensions within the document set, enabling comprehensive coverage of relevant subtopics. Sequential splitting instead fragments related concepts and limits the model's ability to integrate information from multiple perspectives.

In summary, clustering-based pre-construction of a knowledge framework is crucial for generating high-quality structured long-form text, forming a key foundation of the proposed method.

\subsubsection{Computational Efficiency}
\label{subsec:computational_efficiency}
To quantify the latency and computational overhead of ConvergeWriter, we benchmark it against four representative baselines on the same 10 topics. Table~\ref{tab:efficiency_analysis} reports end-to-end wall time and model-token usage together with output length and per-1,000-word measurements; detailed configurations and measurement rules are provided in Appendix~\ref{sec:efficiency_analysis}.

On a per-1,000-word basis, ConvergeWriter requires 92.7 seconds and 21.1K model tokens, placing its computational cost between STORM (34.9 seconds and 16.2K tokens) and OmniThink (181.0 seconds and 105.1K tokens) among multi-stage long-form writing systems. The additional computation reflects ConvergeWriter's evidence-oriented pipeline, including multi-round retrieval and filtering, clustering, cluster summarization, reranking, and section-wise generation, which constructs and exploits an evidence-induced article structure throughout drafting. Together with the clustering ablation and the article-quality, Document Coverage, and claim-level support results reported above, these results characterize a fidelity--efficiency trade-off: ConvergeWriter invests additional computation in evidence-grounded organization, while remaining substantially below OmniThink in both latency and token consumption. A stage-wise cost breakdown is provided in Appendix~\ref{sec:efficiency_analysis}.

\section{Conclusion}
\label{sec:conclusion}

This paper introduces ConvergeWriter, a bottom-up long text generation framework for addressing content credibility challenges in external knowledge base utilization. Its retrieval-first, clustering-structured mechanism constructs a knowledge framework grounded in available evidence via preliminary retrieval and clustering, ensuring generated content remains anchored within knowledge base boundaries. This design guarantees traceability and mitigates hallucinations. Experimental results show that ConvergeWriter achieves performance on par with or superior to state-of-the-art methods across model scales, demonstrating the effectiveness of this knowledge-driven paradigm in closed-knowledge scenarios.

\section{Limitations}

Despite its effectiveness, ConvergeWriter has several limitations. 

First, the framework is designed for knowledge-constrained generation settings where the underlying corpus serves as a relatively reliable knowledge source. Extending the approach to highly open or noisy environments may require additional mechanisms for retrieval quality control and evidence sufficiency assessment. In this direction, iterative RAG methods for question answering, such as FAIR-RAG~\cite{asl2025fairrag}, assess evidence sufficiency for a given query and refine retrieval to address identified evidence gaps. If such query-level refinement were adapted from question answering to section-level evidence assessment in long-form writing, it could complement ConvergeWriter's corpus-level structure induction.

Second, although clustering provides an effective mechanism for organizing retrieved knowledge, the resulting cluster granularity may not always perfectly align with the ideal discourse structure of the target article.

Finally, the retrieval-first pipeline introduces additional computational overhead due to multi-round retrieval, document filtering, clustering, and hierarchical summarization, resulting in higher generation latency compared with top-down approaches.

\section{Ethics Statement}

This work focuses on improving the reliability of long-form document generation by grounding LLM outputs in external knowledge sources. By constraining generation within retrieved evidence, our framework aims to reduce hallucinations and improve factual consistency, which may help mitigate risks associated with misinformation in automated text generation.

Our experiments use publicly available corpora (e.g., Wikipedia) and do not involve personal or sensitive data. A small-scale human evaluation was conducted for research purposes to assess generation quality, where annotators compared model outputs without providing personal information. However, as with other retrieval-augmented generation systems, the quality of generated content still depends on the underlying knowledge sources, and applications in high-stakes domains should rely on trustworthy corpora and appropriate verification mechanisms.

AI-based writing assistants were used to help improve the clarity and language of the manuscript. All technical content was written and verified by the authors.

\section*{Acknowledgments}

This work is funded by the National Natural Science Foundation of China (No.62576085).

\bibliography{custom}
\appendix

\section{Experimental Details}
\subsection{Hyperparameter Settings}
For LLM hyperparameter settings, in text generation tasks, we set \texttt{temperature=0.9} and \texttt{top\_p=0.9} to meet the need for richer and more vivid expressions. Article-generation requests use the API default settings for \texttt{top\_k} and \texttt{min\_p}. During text evaluation, we set \texttt{temperature=0.1} to enhance the rigor and reproducibility of evaluation results.

\subsection{Dataset Subset and Coverage Preprocessing}
\label{sec:evaluation_protocol}
\paragraph{Dataset subset.}
WildSeek contains 100 topics. Because running every generation method under multiple model settings is computationally expensive, our main experiments use the same fixed subset of 60 topics for all methods. Consequently, method comparisons in the main experiments are conducted on matched topics.

\paragraph{Coverage preprocessing.}
We form paragraph-level verification units deterministically. We first split each generated article at Markdown and natural-paragraph boundaries. We then remove headings, empty paragraphs, citation-only fragments, and formatting-only fragments, and merge or discard fragments too short to form independent semantic units. Each remaining paragraph is treated as one verification unit.

\subsection{Claim-Level Evidence Support Analysis}
\label{sec:claim_level_support}
To examine the evidential support of generated content at a finer granularity, we conduct a claim-level evidence support analysis across all five methods. We randomly sample 30 topics and use the same sampled topics for all methods. We process all segments of each article, limiting each segment to 8,000 characters, and use DeepSeek-v4-flash to extract up to three atomic factual claims from each segment. For each claim, we rank the saved documents associated with the corresponding method--topic pair using BM25 and provide the top five documents, each limited to 12,000 characters, to DeepSeek-v4-flash for verification. We define the claim support rate as the number of supported claims divided by the number of verified claims. The results are shown in Table~\ref{tab:claim_level_support}.

\begin{table}[H]
\centering
\small
\setlength{\tabcolsep}{3.5pt}
\begin{tabular}{lrrr}
\toprule
Method & Claims & Supported & Support Rate \\
\midrule
Direct RAG & 815 & 582 & \underline{0.714} \\
Two-Stage RAG & 2,603 & 148 & 0.057 \\
STORM & 1,590 & 803 & 0.505 \\
OmniThink & 920 & 472 & 0.513 \\
ConvergeWriter & 818 & 589 & \textbf{0.720} \\
\bottomrule
\end{tabular}
\caption{Claim-level evidence support results on the same 30 randomly sampled topics. Claims and Supported denote the numbers of verified and evidence-supported claims, respectively. Bold and underlined values denote the best and second-best support rates, respectively.}
\label{tab:claim_level_support}
\end{table}

As shown in Table~\ref{tab:claim_level_support}, ConvergeWriter achieves the highest claim support rate of 0.720, providing further claim-level support for the advantage of its bottom-up, data-driven strategy in terms of verifiability and credibility.

\subsection{Parallel Evaluation with a Heterogeneous LLM Judge}
\label{sec:heterogeneous_llm_judge}
We conduct a parallel evaluation of the same articles reported in the Qwen3-32B-with-reasoning block of Table~\ref{tab:results} using DeepSeek-v4-flash as a heterogeneous LLM judge. The generated articles, five methods, and fixed set of 60 common topics remain the same across the Qwen and DeepSeek evaluations. Method names are hidden during evaluation, and the same four rubric dimensions---\textit{Relevance}, \textit{Breadth}, \textit{Depth}, and \textit{Novelty}---are used. The evaluation results are presented in Table~\ref{tab:deepseek_eval}.

\begin{table}[H]
\centering
\small
\resizebox{\columnwidth}{!}{%
\begin{tabular}{lccccc}
\toprule
Method & Relevance & Breadth & Depth & Novelty & AVG \\
\midrule
Direct RAG & \textbf{4.900} & 4.133 & 4.017 & \underline{2.700} & 3.938 \\
Two-Stage RAG & 3.233 & 3.517 & 3.483 & 2.200 & 3.108 \\
STORM & 4.767 & \underline{4.400} & \underline{4.317} & 2.633 & \underline{4.029} \\
OmniThink & 3.967 & 3.617 & 3.300 & 2.600 & 3.371 \\
ConvergeWriter & \underline{4.817} & \textbf{4.483} & \textbf{4.433} & \textbf{3.000} & \textbf{4.183} \\
\bottomrule
\end{tabular}
}
\caption{Evaluation on the same fixed 60-topic article set using DeepSeek-v4-flash as a heterogeneous LLM judge. Bold and underlined values denote the best and second-best results, respectively.}
\label{tab:deepseek_eval}
\end{table}

Overall, ConvergeWriter ranks first in three dimensions and in average score under the DeepSeek judge, broadly aligning with the original Qwen evaluation and providing complementary cross-judge evidence.

\subsection{Human--LLM Evaluation Consistency}
\label{sec:llm_human_consistency}
To compare the LLM-based evaluation used in our main experiments with human judgments, we conduct a human evaluation study. We randomly sample 32 article pairs from the test set, where each pair contains two articles generated by different methods under the same topic. Eight researchers with computer science backgrounds independently perform blind pairwise comparisons across the four evaluation dimensions: \textit{Relevance}, \textit{Breadth}, \textit{Depth}, and \textit{Novelty}. For each dimension, annotators indicate whether Article A is better, Article B is better, or the two are tied.

\begin{table}[H]
\centering
\small
\begin{tabular}{lcccc}
\toprule
Metric & Agreement & Kendall $\tau$ & Cohen $\kappa$ & N \\
\midrule
Relevance & 0.60 & 0.46 & 0.31 & 10 \\
Breadth & 0.71 & 0.49 & 0.46 & 21 \\
Depth & 0.78 & 0.64 & 0.58 & 18 \\
Novelty & 0.64 & 0.32 & 0.30 & 22 \\
\midrule
Overall & 0.69 & 0.46 & 0.42 & 71 \\
\bottomrule
\end{tabular}
\caption{Consistency between human evaluation and LLM-based scoring under strict filtering (excluding LLM tie cases).}
\label{tab:human_llm_consistency}
\end{table}

Since the LLM evaluator assigns discrete scores (1--5), ties can occur when converting scores into pairwise preferences. For the consistency analysis, we apply strict filtering by retaining only dimension-level comparisons in which the two articles receive different LLM scores. The resulting agreement statistics are shown in Table~\ref{tab:human_llm_consistency}. Across the resulting 71 comparisons, the overall agreement rate is 0.69, with Kendall's $\tau$ of 0.46 and Cohen's $\kappa$ of 0.42. These results indicate a degree of consistency between LLM-based scores and human judgments under this comparison setting.

\subsection{Topic-Level Paired Significance Tests}
\label{sec:paired_significance}
For the Qwen3-32B-with-reasoning setting, we conduct topic-level paired tests on the same fixed set of 60 common topics. For each topic, we calculate the difference between ConvergeWriter's average score across the four rubric dimensions and that of each baseline under the original Qwen judge. We report the mean difference, a 95\% confidence interval estimated with 10,000 paired bootstrap resamples, wins/losses/ties, and a two-sided sign-test $p$-value computed over non-tied topic pairs.

\begin{table*}[t]
\centering
\small
\setlength{\tabcolsep}{5.5pt}
\begin{tabularx}{\textwidth}{l *{5}{>{\centering\arraybackslash}X}}
\toprule
Method & AVG & $\Delta$ & 95\% Bootstrap CI & Wins/Losses/Ties & Sign-test $p$ \\
\midrule
\rowcolor{gray!22}
ConvergeWriter & 4.858 & -- & -- & -- & -- \\
Direct RAG & 4.625 & +0.233 & $[+0.113, +0.396]$ & 41/4/15 & $1.0 \times 10^{-8}$ \\
Two-Stage RAG & 4.237 & +0.621 & $[+0.325, +0.954]$ & 32/8/20 & 0.0001822 \\
STORM & 4.533 & +0.325 & $[+0.125, +0.575]$ & 32/9/19 & 0.0004309 \\
OmniThink & 4.329 & +0.529 & $[+0.354, +0.717]$ & 42/9/9 & $3.39 \times 10^{-6}$ \\
\bottomrule
\end{tabularx}
\caption{Topic-level paired significance tests for the average score across the four rubric dimensions under the original Qwen judge. $\Delta$ denotes ConvergeWriter minus the listed baseline; dashes indicate that the paired-comparison statistics are not applicable to the ConvergeWriter reference row.}
\label{tab:paired_significance}
\end{table*}

As shown in Table~\ref{tab:paired_significance}, all four comparisons yield positive mean differences. The paired bootstrap confidence intervals exclude zero, and the two-sided sign-test $p$-values are below 0.001 in all cases. These results indicate that, under the original Qwen judge, ConvergeWriter's average score across the four rubric dimensions is significantly higher than that of each baseline in the topic-level paired comparisons.

\subsection{Sensitivity to the Number of Retrieved Documents}
\label{sec:coverage_sensitivity}
To examine the effect of the number of retrieved documents on Document Coverage, we use Qwen3-8B with thinking disabled as the verifier and evaluate $k\in\{1,2,3,5\}$ on the same 60 topics for ConvergeWriter, Direct RAG, STORM, and OmniThink. Across different values of $k$, we use the same paragraph segmentation, BM25 retrieval, and strict support-verification procedure, without truncating paragraphs or retrieved documents. The results are shown in Table~\ref{tab:coverage_sensitivity}.

\begin{table}[H]
\centering
\small
\begin{tabular}{lcccc}
\toprule
Method & $k=1$ & $k=2$ & $k=3$ & $k=5$ \\
\midrule
Direct RAG & 27.88 & 34.85 & 42.42 & 40.61 \\
STORM & 18.04 & 24.05 & 26.73 & 31.82 \\
OmniThink & 28.50 & 32.05 & 35.10 & 32.58 \\
ConvergeWriter & 33.45 & 40.27 & 52.22 & 45.73 \\
\bottomrule
\end{tabular}
\caption{Document Coverage (\%) under different numbers of retrieved documents.}
\label{tab:coverage_sensitivity}
\end{table}

As shown in Table~\ref{tab:coverage_sensitivity}, ConvergeWriter achieves the highest Coverage at every tested value of $k$. Meanwhile, increasing $k$ does not yield consistent monotonic improvements. Considering both the evaluation results and computational cost, we use $k=2$ for the Document Coverage evaluation to balance evidence availability and verification cost: compared with $k=1$, it provides more evidence while limiting the additional context noise and context-length risks associated with larger values of $k$.

\subsection{Qualitative Case Study of Generated Outputs}
\label{sec:qualitative_case_study}

To provide a concrete qualitative view of the generated outputs, we examine ConvergeWriter and STORM on the topic ``Artificial Intelligence and Copyright Dilemma.'' On this topic, ConvergeWriter received an average score of 4.5 from the DeepSeek-v4-flash judge across the four evaluation dimensions, and the claim-level verifier classified 30 of 33 extracted claims as supported by its saved retrieved evidence (0.909); the corresponding results for STORM were 4.0 and 37 of 59 (0.627). Table~\ref{tab:qualitative_case_study} compares their outline and output patterns with the assessments against their respective saved evidence pools.

\begin{table*}[!t]
\centering
\small
\setlength{\tabcolsep}{4pt}
\renewcommand{\arraystretch}{1.15}
\begin{tabularx}{\textwidth}{>{\raggedright\arraybackslash}p{0.13\textwidth} >{\raggedright\arraybackslash}X >{\raggedright\arraybackslash}p{0.23\textwidth} >{\raggedright\arraybackslash}X}
\toprule
\textbf{Case Aspect} & \textbf{STORM: Top-Down Outline and Output} & \textbf{Evidence-Pool Assessment} & \textbf{ConvergeWriter: Retrieval-First Outline and Output} \\
\midrule
Regulatory requirements & The outline includes ``AI Act transparency/accountability provisions,'' followed by the claim that the EU AI Act mandates documentation of training data for high-risk systems. & The saved retrieved documents do not explicitly mention this training-data documentation requirement. & The article organizes this discussion under broader sections on ``AI and Generative Technologies'' and ``Regulatory and Ethical Challenges,'' without making this specific claim. \\
\midrule
U.S. copyright decisions & The outline includes ``U.S. Copyright Office guidelines on AI-generated content,'' and the article states that the U.S. Copyright Office rejected an AI-generated art portfolio in 2023. & The saved retrieved documents do not mention this specific 2023 rejection. & The article discusses general issues of authorship and ownership without referring to this specific institutional event. \\
\midrule
Legal and policy implications & The outline includes ``Future Directions,'' ``global harmonization,'' and ``liability for AI-generated infringement,'' and the article makes claims about future disputes and rights allocation for AI-generated works. & Several future-impact and policy-implication claims in these sections are not explicitly supported by the saved retrieved documents. & The article organizes the discussion around generative AI, copyright frameworks, and moral rights and authorship drawn from its evidence clusters. \\
\midrule
Case-level summary & The generated article presents a broad legal and policy discussion. & Eighteen unsupported claims fall into at least one of the analyzed categories: ethical issues, future impacts, or policy implications. & The article uses a broader structure centered on retrieved evidence clusters. \\
\bottomrule
\end{tabularx}
\caption{Qualitative comparison of the outlines and generated outputs of ConvergeWriter and STORM on ``Artificial Intelligence and Copyright Dilemma.'' Evidence-pool assessments indicate whether the displayed STORM statements are explicitly supported by STORM's saved retrieved evidence.}
\label{tab:qualitative_case_study}
\vspace{0.8\baselineskip}

\centering
\small
\setlength{\tabcolsep}{5pt}
\renewcommand{\arraystretch}{1.12}
\begin{tabularx}{\textwidth}{>{\raggedright\arraybackslash}X r r r r}
\toprule
\textbf{Pipeline phase / major timed component} & \textbf{Mean wall (s)} & \textbf{Share} & \textbf{Model tokens} & \textbf{LM calls} \\
\midrule
Phase I & 103.6 & 35.8\% & 41,852.3 & 154.9 \\
\quad Retrieval and filtering & 11.1 & -- & -- & -- \\
\quad Expansion-keyword generation & 4.4 & -- & -- & -- \\
\quad Embedding and clustering & 10.9 & -- & -- & -- \\
\quad Cluster summarization & 52.6 & -- & -- & -- \\
\addlinespace
Phase II & 185.5 & 64.2\% & 24,594.3 & 5.9 \\
\quad Reranking & 87.2 & -- & -- & -- \\
\bottomrule
\end{tabularx}
\caption{Phase-level and major-component computational breakdown of ConvergeWriter on the auxiliary 10-topic efficiency benchmark. Phase I comprises relevance-expanding retrieval, knowledge structurization, and structured knowledge mapping-based outline generation; Phase II comprises outline--knowledge cluster joint-driven section-wise retrieval and article generation. Only the principal timed components are shown; minor pipeline operations are omitted.}
\label{tab:efficiency_breakdown}
\end{table*}

\subsection{Computational Efficiency Details}
\label{sec:efficiency_analysis}

We conduct the auxiliary efficiency benchmark on the same set of 10 topics for all five methods. To control variation arising from model context limits, online Wikipedia retrieval, and external service concurrency, all methods use DeepSeek-v4-flash for LLM calls and retrieve from a dense index of the preprocessed English Wikipedia corpus released for HotpotQA, derived from the October 1, 2017 Wikipedia dump and hosted on a local server. Wall time denotes the end-to-end elapsed time for processing a single topic, and model-token usage is the sum of prompt and completion tokens. Seconds and model tokens per 1,000 words are computed separately for each topic and then averaged across the 10 topics. Each method retains the native internal parallelism and output-length configuration used in its benchmark implementation.

The phase-level breakdown shows that 35.8\% of ConvergeWriter's end-to-end time is devoted to retrieval, knowledge structurization, and outline generation, while 64.2\% is devoted to article generation. Among the major timed components, reranking and cluster summarization have the largest wall times. This distribution complements the fidelity--efficiency trade-off discussed in Section~\ref{subsec:computational_efficiency}.

\subsection{Details of Further Analysis}
\begin{table}[H]
    \centering
    \resizebox{\columnwidth}{!}{%
        \begin{tabular}{l c c c c c}
            \toprule
            Methods & Relevance & Breadth & Depth & Novelty & AVG \\
            \midrule
            ConvergeWriter & 4.97 & 4.95 & 4.93 & 4.58 & 4.86 \\
            w/o Clustering & 4.93 & 4.87 & 4.93 & 3.60 & 4.58 \\
            \bottomrule
        \end{tabular}
    }
    \caption{Ablation results with and without document clustering. Bold values indicate the better result in each column.}
    \label{tab:ablation_results}
\end{table}

In Section~\ref{sec:further_analysis}, we evaluated the performance gap before and after removing the clustering module in ConvergeWriter. Specific metric scores are shown in Table~\ref{tab:ablation_results}.

\section{Prompts Used}

\begin{figure}[H]
\centering
\includegraphics[width=0.9\columnwidth]{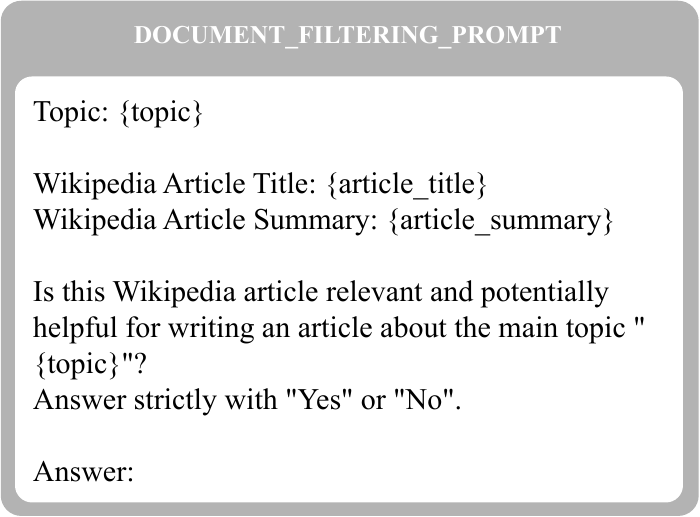} 
\caption{Prompt for document filtering.}
\label{fig:prompt_document_filtering}
\end{figure}

\begin{figure}[H]
\centering
\includegraphics[width=0.9\columnwidth]{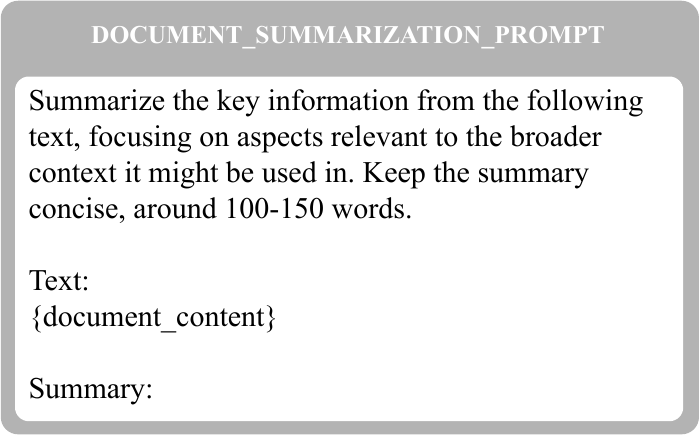} 
\caption{Prompt for document summarization.}
\label{fig:prompt_document_summarization}
\end{figure}

\begin{figure}[b]
\centering
\includegraphics[width=0.9\columnwidth]{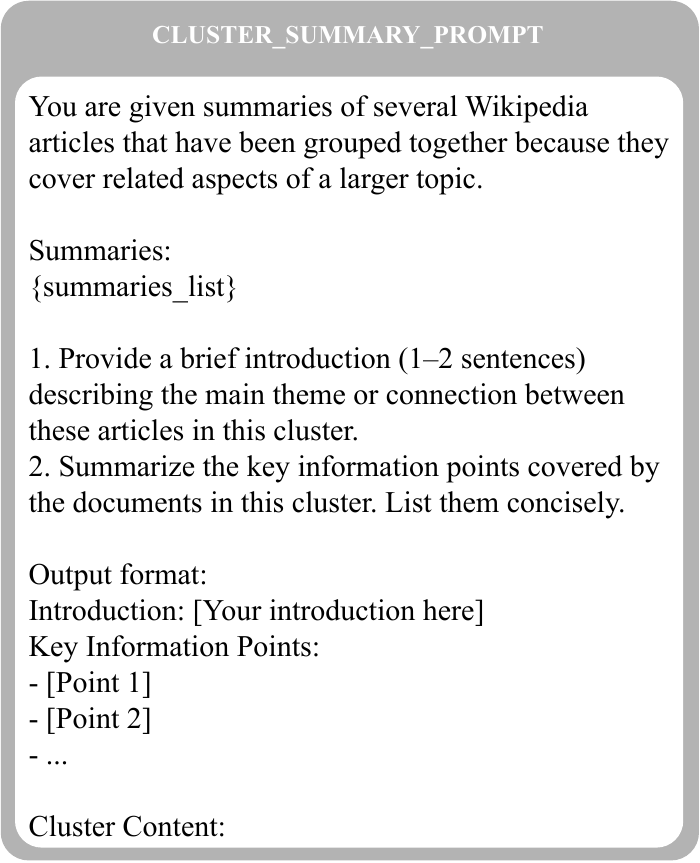} 
\caption{Prompt for cluster summarization.}
\label{fig:prompt_cluster_summarization}
\end{figure}

\begin{figure*}[t]
\centering
\includegraphics[width=0.95\textwidth]{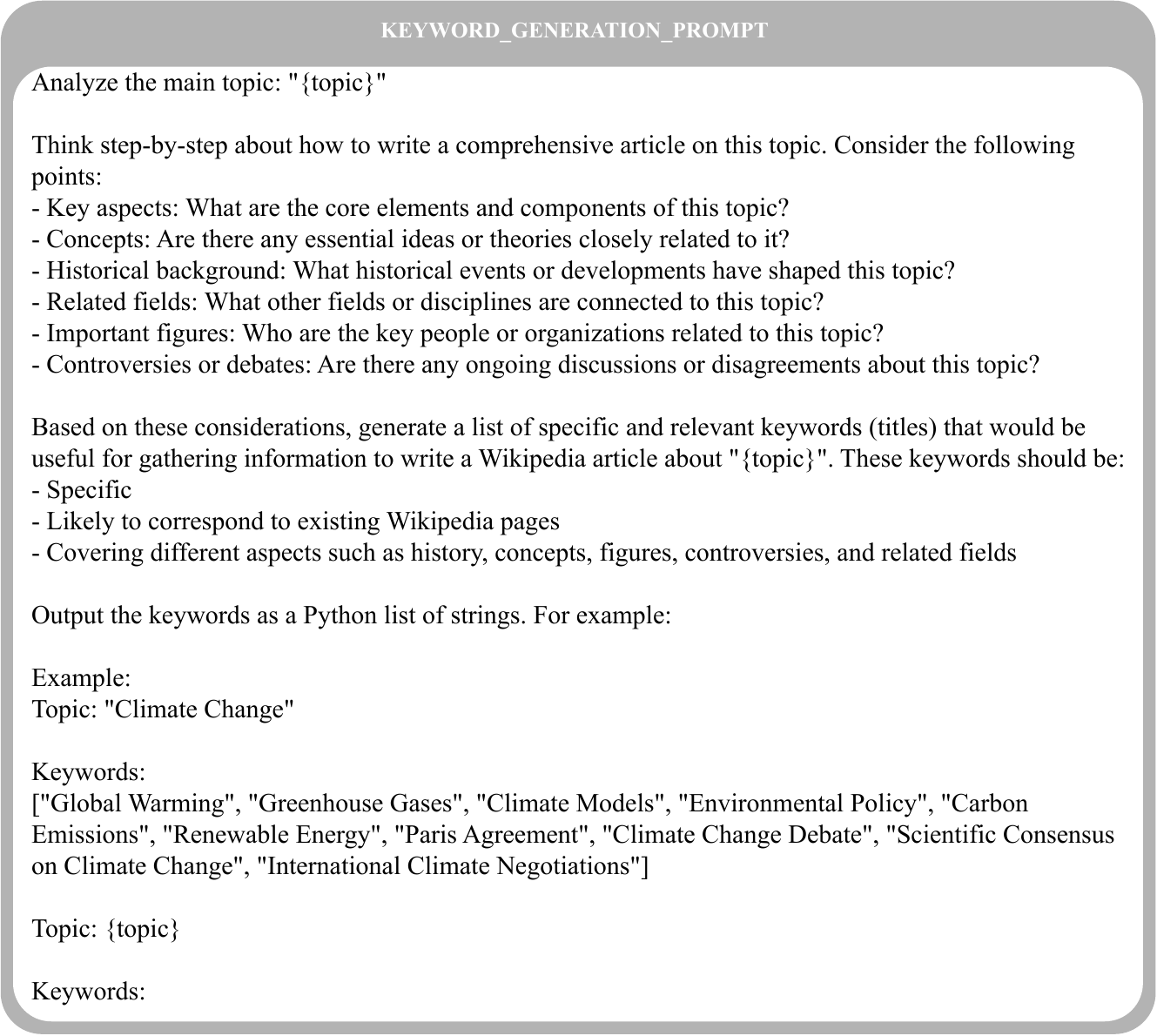} 
\caption{Prompt for keyword generation.}
\label{fig:prompt_keyword_generation}
\end{figure*}

\begin{figure*}[t]
\centering
\includegraphics[width=0.95\textwidth]{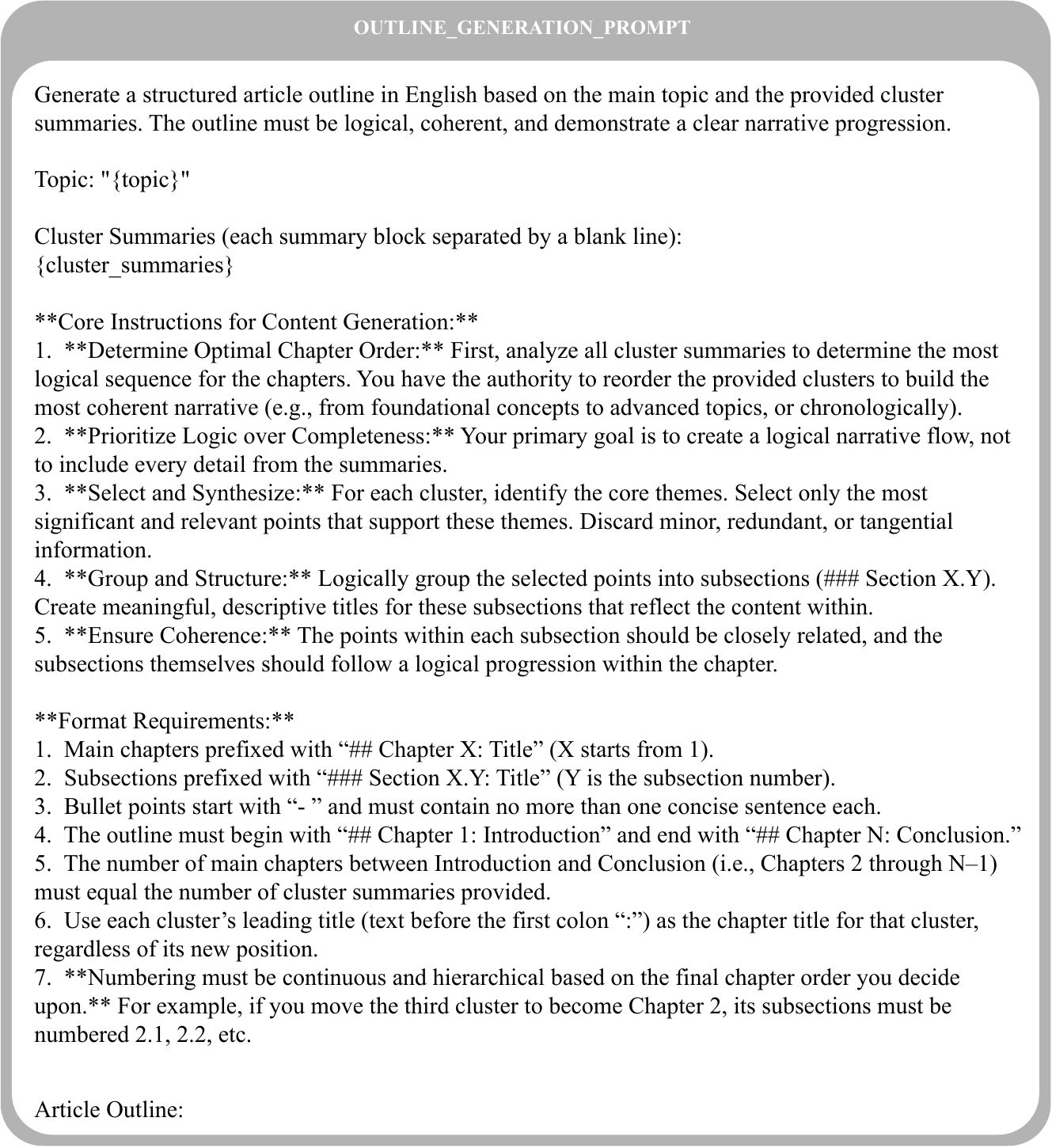} 
\caption{Prompt for outline generation.}
\label{fig:prompt_outline_generation}
\end{figure*}

\begin{figure*}[t]
\centering
\includegraphics[width=0.95\textwidth]{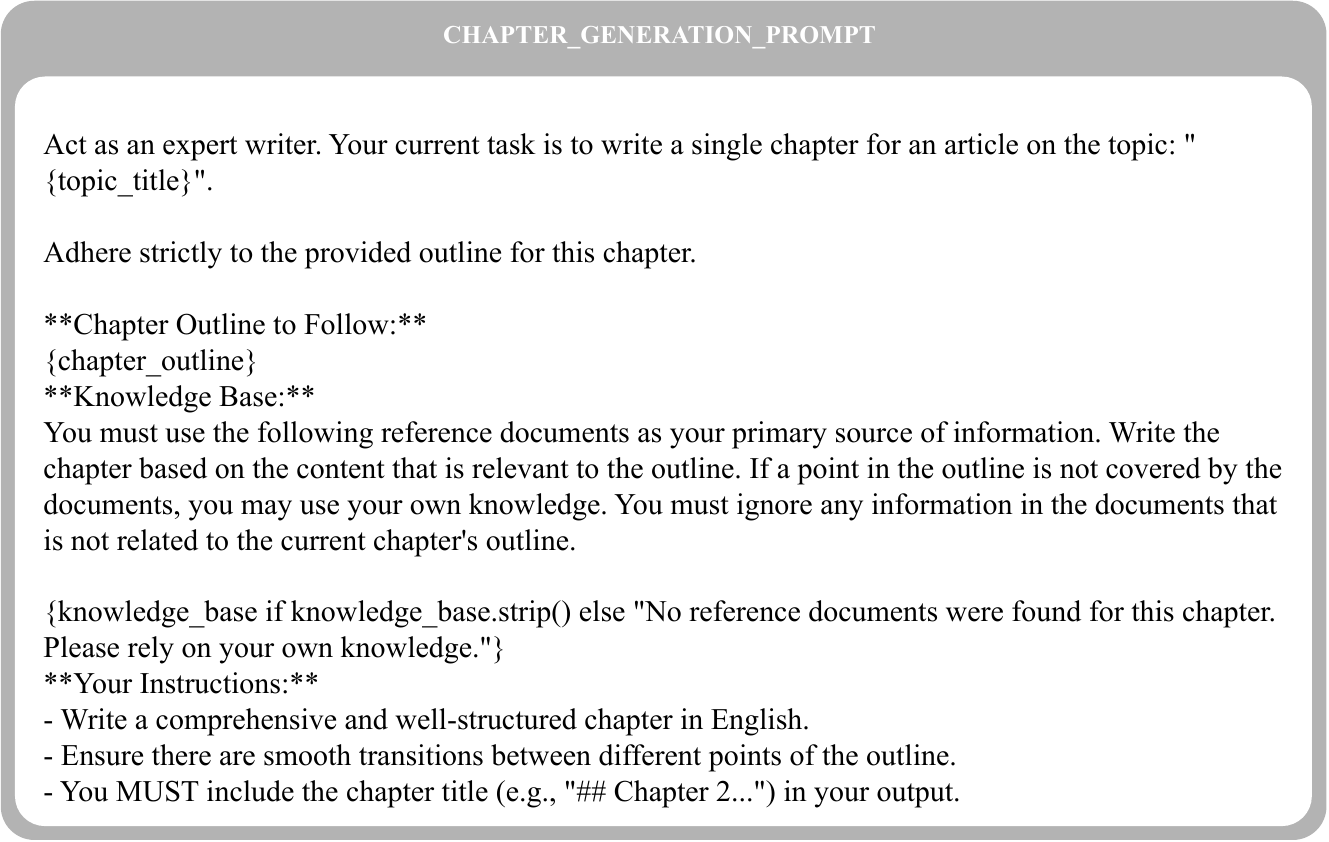} 
\caption{Prompt for chapter generation.}
\label{fig:prompt_chapter_generation}
\end{figure*}

\begin{figure*}[t]
\centering
\includegraphics[width=0.95\textwidth]{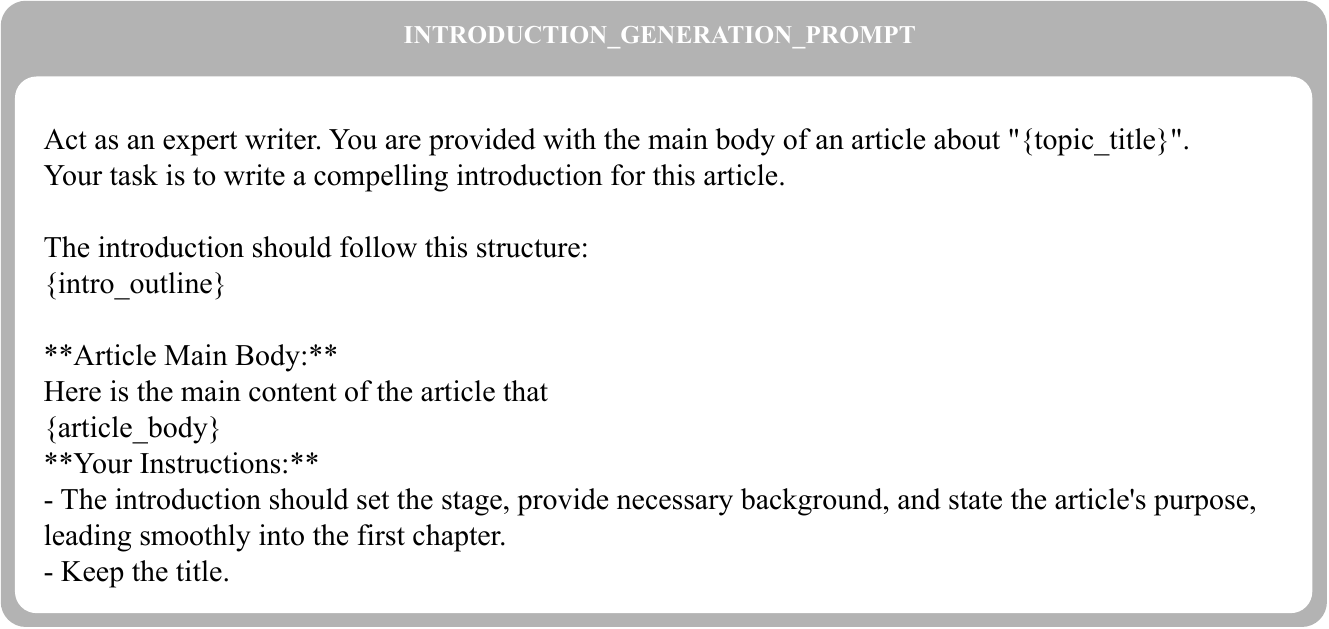} 
\caption{Prompt for introduction generation.}
\label{fig:prompt_introduction_generation}
\end{figure*}

\begin{figure*}[t]
\centering
\includegraphics[width=0.95\textwidth]{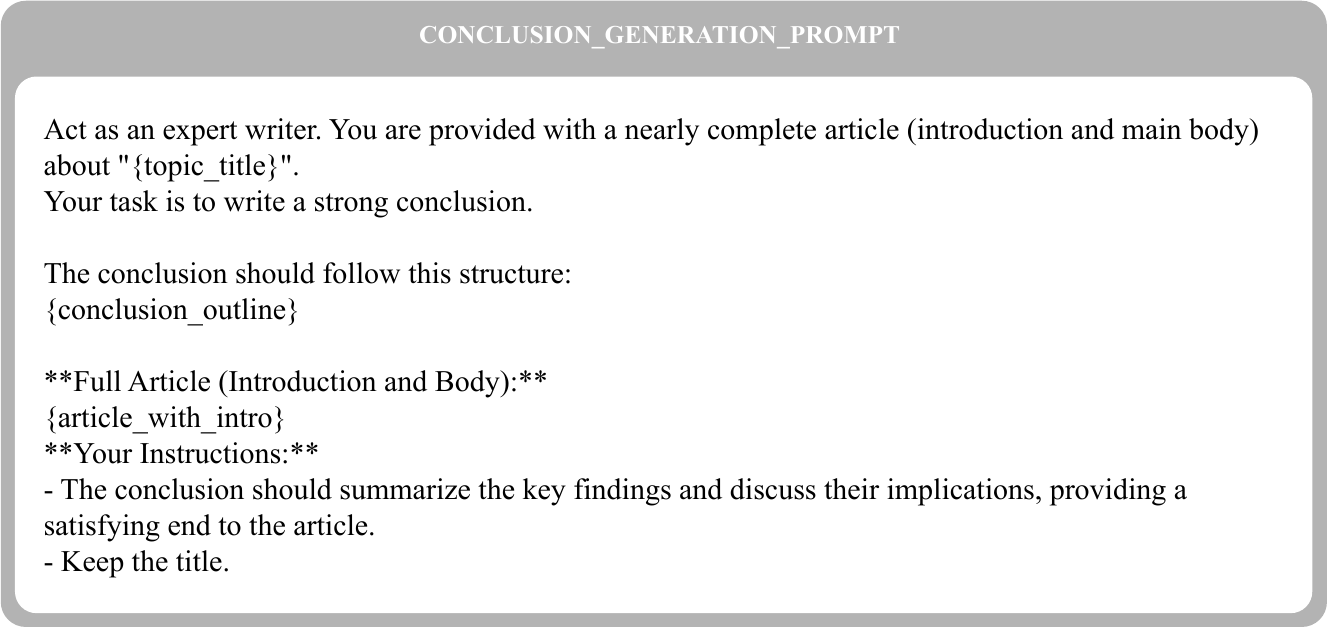} 
\caption{Prompt for conclusion generation.}
\label{fig:prompt_conclusion_generation}
\end{figure*}

\begin{figure*}[t]
\centering
\includegraphics[width=0.95\textwidth]{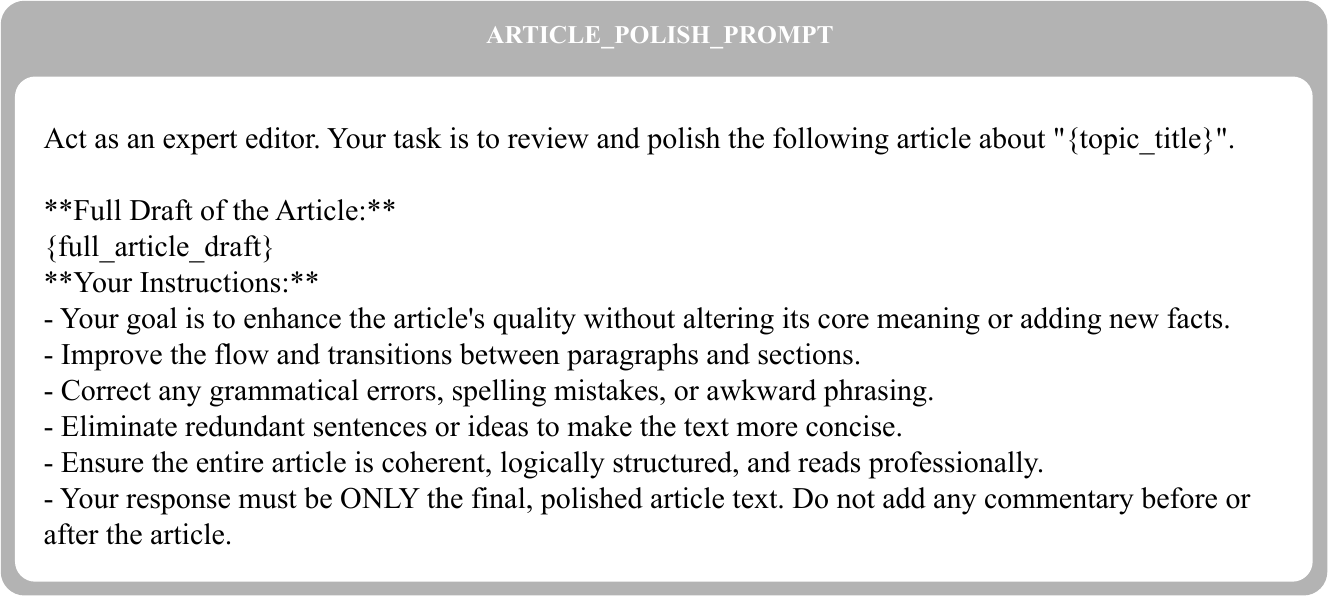} 
\caption{Prompt for final article polishing.}
\label{fig:prompt_article_polishing}
\end{figure*}

\end{document}